\documentclass[a4paper]{article}
\usepackage{url}
\usepackage[english]{babel}
\usepackage[utf8]{inputenc}
\usepackage{amsmath}
\usepackage{graphicx}
\usepackage{color}
\usepackage{float}
\usepackage{listings}
\usepackage{tikz}
\usetikzlibrary{shapes.geometric, arrows}
\definecolor{keywords}{RGB}{255,0,90}
\definecolor{comments}{RGB}{0,0,113}
\definecolor{red}{RGB}{160,0,0}
\definecolor{green}{RGB}{0,150,0}
\definecolor{codegreen}{rgb}{0,0.6,0}
\definecolor{codegray}{rgb}{0.5,0.5,0.5}
\definecolor{codepurple}{rgb}{0.58,0,0.82}
\definecolor{backcolour}{rgb}{0.95,0.95,0.92}
\definecolor{brown}{rgb}{0.59, 0.29, 0.0}
\definecolor{beaublue}{rgb}{0.74, 0.83, 0.9}
\definecolor{orange}{rgb}{1.0, 0.5, 0.0}
\definecolor{darkslategray}{rgb}{0.18, 0.31, 0.31}
\definecolor{deepblue}{rgb}{0,0,0.5}
\definecolor{deepred}{rgb}{0.6,0,0}
\definecolor{deepgreen}{rgb}{0,0.5,0}
\definecolor{auburn}{rgb}{0.43, 0.21, 0.1}
\definecolor{bistre}{rgb}{0.24, 0.17, 0.12}
\definecolor{babyblue}{rgb}{0.54, 0.81, 0.94}
\definecolor{ballblue}{rgb}{0.13, 0.67, 0.8}
\lstdefinestyle{myMatlabstyle}{
	language=Matlab,
	backgroundcolor=\color{white},   
	commentstyle=\color{deepgreen},
	keywordstyle=\color{black},
	identifierstyle=\color{black},
	numberstyle=\tiny\color{codegray},
	stringstyle=\color{purple},
	basicstyle=\footnotesize,
	breakatwhitespace=false,         
	breaklines=true,                 
	captionpos=b,                    
	keepspaces=true,                 
	numbers=left,                    
	numbersep=5pt,                  
	showspaces=false,                
	showstringspaces=false,
	showtabs=false,                  
	tabsize=2
}
\lstdefinestyle{myPythonstyle}{
	language=Python, 
	basicstyle=\ttfamily\small, 
	keywordstyle=\color{blue},
	commentstyle=\color{green},
	stringstyle=\color{red},
	showstringspaces=false,
	identifierstyle=\color{black},
}
\usepackage[colorinlistoftodos]{todonotes}
\usepackage[scale=0.75]{geometry}
	\title{Estimating speech from lip dynamics}

\author{Jithin George, Ronan Keane, Conor Zellmer}

\date{\today}

\begin{document}
\maketitle

\begin{abstract}
The goal of this project is to develop a limited lip reading algorithm for a subset of the English language. We consider a scenario in which no audio information is available. The raw video is processed and the position of the lips in each frame is extracted. We then prepare the lip data for processing and classify the lips into visemes and phonemes. Hidden Markov Models are used to predict the words the speaker is saying based on the sequences of classified phonemes and visemes. The GRID audiovisual sentence corpus \cite{key-10}\cite{key-11} database is used for our study. 
\end{abstract}

\section{Introduction and Overview}
\label{sec:introduction}

The capacity of machines to learn and understand this world grows daily. Many of their feats surpass human limits. Every day, they reveal incredible abilities that were thought impossible before. One such example is \cite{key-1} where audio could be retrieved from a video by analyzing minute vibrations of objects in it ranging from a glass of water to a bag of chips. We have a much more humble goal of extracting the spoken words from the visuals in a video.

This is an attempt to use classification algorithms and Hidden Markov Models to perform visual speech recognition.  Videos from the GRID Audiovisual Sentence Corpus database \cite{key-10}\cite{key-11}  are used in the project. The GRID database contains a set of 1000 videos of a single speaker. Each video is approximately 3 seconds in length, and contains a man speaking a collection of words chosen to encompass each sound possible in spoken English. The database includes transcripts of each video, with the locations of each word in the video by frame numbers.  

This report details the extraction of lip contours from each video, the identification of phonemes being spoken using classification methods, and prediction of speech using Hidden Markov Models (HMM).  The lip contours for each video frame are extracted using active contour masks, dynamic mode decomposition, edge detection and color classification. Then we use classification algorithms like naive Bayes and k-nearest neighbors to map the contours to phonemes/visemes.  Finally we use hidden Markov models to analyze sequences of classified phonemes/visemes to predict what words they are.

\section{Theoretical Background}
\label{sec:theory}
\subsection{Feature Extraction}
Lip reading is a complicated task and there are no "go-to" algorithms for detecting and tracking the position of an individual's lips. We can use Matlab's built in active contour and edge detection to hope to do background/foreground separation on the video. We can also use DMD to do background/foreground separation. The assumption is the speaker's face is stationary enough that it is possible to detect the lips as the foreground. In practice, this is not the case. We can also classify the pixel colors of the video frames and segment the lips based on the idea that the speaker's lip color is different from their skin color. For color classification, we can project the pixel color into the LAB color space to achieve better classification results. In either of these different strategies we also need to isolate the general mouth region of the speaker so that we will not also detect eyes, nose, etc.
\subsection{Phonemes and Visemes}
Phonemes are the smallest identifiable sound unit in the sound system of a language. $^{[6]}$  According to Zhong, et al, phonemes are ``basic sound segments that carry linguistic distinction.''$^{[7]}$ In theory, visemes are the analogous basic units in the visual domain.  However, there is no agreement on a common definition for visemes in practice$^{[8]}$.  In audio speech recognition, phonemes are detected and used to reconstruct speech.  In visual speech recognition, only visemes can be detected.  Phonemes for the basis for a spoken language, and hence automated lipreading typically employs a mapping from phonemes to visemes.  Many such mapping can be found in the literature, but all suffer from the issue of there being more phonemes than visemes, resulting in a many-to-one map.$^{[8]}$  For instance, this project uses 37 phonemes and 11 visemes. See Table (\ref{tab:pv}) for the phoneme to viseme map used in this project.  

\begin{table}[!ht]
\center
\caption{Phoneme to Viseme Map from Lee and York, 2000, via [8].}
\begin{tabular}{l l l}
\hline\hline\\
Viseme Number & Viseme Label & Associated Phonemes\\
 \hline
1   & P      &  b p m \\ 
2   & T      &  d t s z th dh\\
3   & K      &   g k n l y hh\\
4   & CH   &   jh ch\\
5   & F      &   f v\\
6   & W     &   r w\\
7   & IY     &   iy ih\\
8   & EH   &   eh ey ae\\
9   & AA    &   aa aw ay ah\\
10   & A0  &   ao oy ow\\
11 & UH   &   uh uw\\
\end{tabular}
\label{tab:pv}
\end{table}

\subsection{Hidden Markov Models}

A Markov model involves the transition of a particular state to other states based on transition probabilities. A future state is only depends on the current state and not the states before it. Now, consider that at every state, there would be real world observations. These observations are controlled by the emission probabilities at each state.

For example, if we were to represent the ever changing weather, the states would be sunny, rainy or snowing and the observations would be summer clothes, rain boots or snow shoes. We can see that the emission probabilities for each observation is different depending on the state. To be more clear, the emission probabilities depend on the states.

We look at Hidden Markov Models(HMM). We decide that this is a relevant model because the words spoken are those defined by language and thus occur in specific pattern and not randomly. For example, given the first letter of word 'k', the probability that the next letter is a vowel is much higher than it being a consonant. A machine learning algorithm without this would be as inefficient as the initial Enigma machine in the movie "The Imitation Game". HMMs are very popular in the fields of speech \cite{key-2} and gesture recognition \cite{key-4} \cite{key-5}.

Although HMMs have fascinating problems related to evaluation and decoding, our interests are in learning. We have a sequence of observations and our aim is to estimate the transition and observation matrices that created that. The Baum-Welch algorithm \cite{key-3} gives us that and we train a transition and observation matrix for each word by using sequences corresponding to different instances of that word. In our weather example, we could take the sequences of weather in a few city and train an HMM to find which city produces a test sequence of weather. A month of sunshine is unlikely to be Seattle but California would be a better guess.

So, given the features from the videos, we find the states. The states are the units of words, here chosen to be phonemes. 
\section{Implementation and Development}

\subsection{Feature extraction}

\subsubsection{Obtaining the initial mask}
The first step is to determine the general location of the mouth in each frame. Matlab has a built in facial recognition package ``Cascade Vision Detector'' which uses the Viola-Jones algorithm to detect the facial features of people. We use the Cascade Mouth detector to determine the location of the face in each frame of each movie. This algorithm is not perfect however and we often detect the eyes or chin of the subject in question. We filter out these false detections by looking at the position of the mouth. The speaker is in generally the same location in each movie so we know if a detected mouth position is too low or too high we can throw that region out. We refer to this detected mouth region as the \textit{initial mask}. We need a good initial mask in order to properly filter our results. Without knowing the general mouth region our algorithms will act on the whole face, but we consider only the lips for our project.Therefore, really our objective is to separate the lips from the mouth region.\par

\subsubsection{Active contour}
Next we convert the image to grayscale. On the grayscale image we can apply Matlab's built in active contour and edge detection, as well as the DMD algorithm we develop in homework 4. We can experiment with the different segmentation types for Matlab's active contour and edge detection. For active contour, Chan-Vese gives poor results with irregular edges. Edge method for active contour gives smoother regions but irregular shapes. Active contour takes many iterations to properly converge and gives poor results. From the figures of active contour, we see the inconsistent lip region detected. Because we are trying to determine the changes in lip shape over time, active contour is not suitable because the results it produces vary too much. Additionally, because it is extremely slow, active contour is not ideal for our purposes. We need to process a large number of videos so a fast algorithm is preferable. If active contour was more computationally efficient it might be possible to use a large number of iterations for each frame in order to obtain a consistent result.  \par
\begin{figure}[!ht]
	\includegraphics[width=1\textwidth, height=0.75\textwidth]{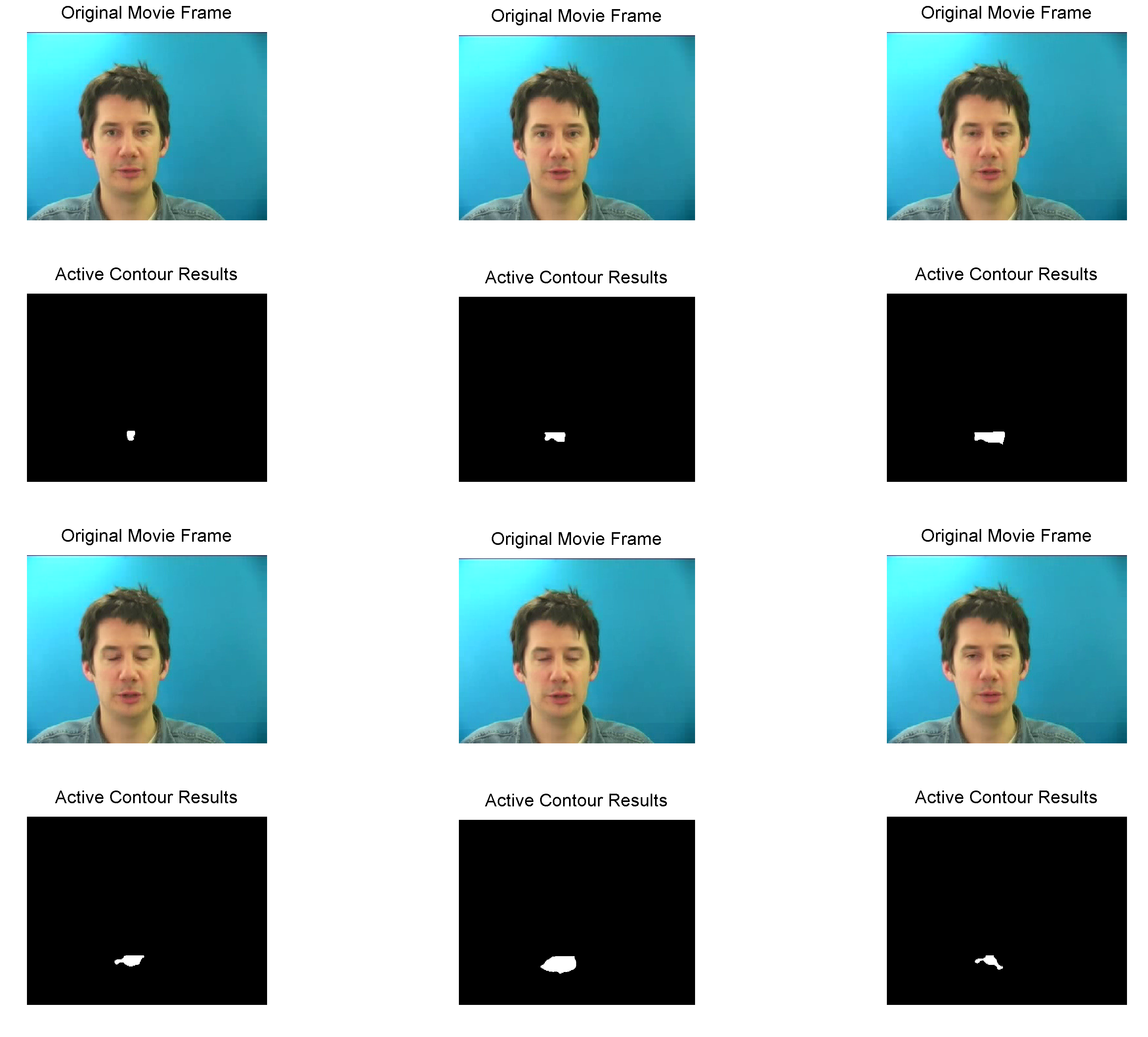}
	\caption{Active Contour Results for selected frames. Active contour provides an inconsistent lip profile because it never properly converges.}
\end{figure}
\subsubsection{Canny edge detection}
Matlab's built in edge detection has many different methods which produce different results. We consider using Canny, Prewitt and Sobel edge detection.Canny typically finds the ``strong'' edges whereas Sobel and Prewitt tend to pick out more detail in the image. We are really only interested in the lips, and ideally want only 1 contour, so we decide to use Canny edge detection. We want to detect the outside and inside edge of the lips, but all methods typically detect  the inside edge of the lip. When using Matlab's ``edge'', it is important to consider the threshold. Too low a threshold value will produce other features of the face and possibly noise. Too high a threshold and we will not pick out the entire lip. Therefore we decide to start at a high threshold value, and if we do not find a large enough edge, we lower the threshold and do the detection again. Typically we only need the lower the threshold once or twice. We set a minimum number of detected pixels in the edge to decide whether or not to automatically lower the threshold. Although this may seem expensive, in practice we only need to lower the threshold for certain frames, and so most of the time we only need to do the algorithm once.  Additionally, edge is much faster than active contour, so this method is still much faster than active contour, and gives more consistent results in well. Overall the Canny edge detection finds a good edge for the lips but it oftentimes only picks out part of the mouth or sometimes picks out other features around the mouth as well, like the chin or space between the nose and mouth.\par
\begin{figure}[!ht]
	\includegraphics[width=1\textwidth, height=0.75\textwidth]{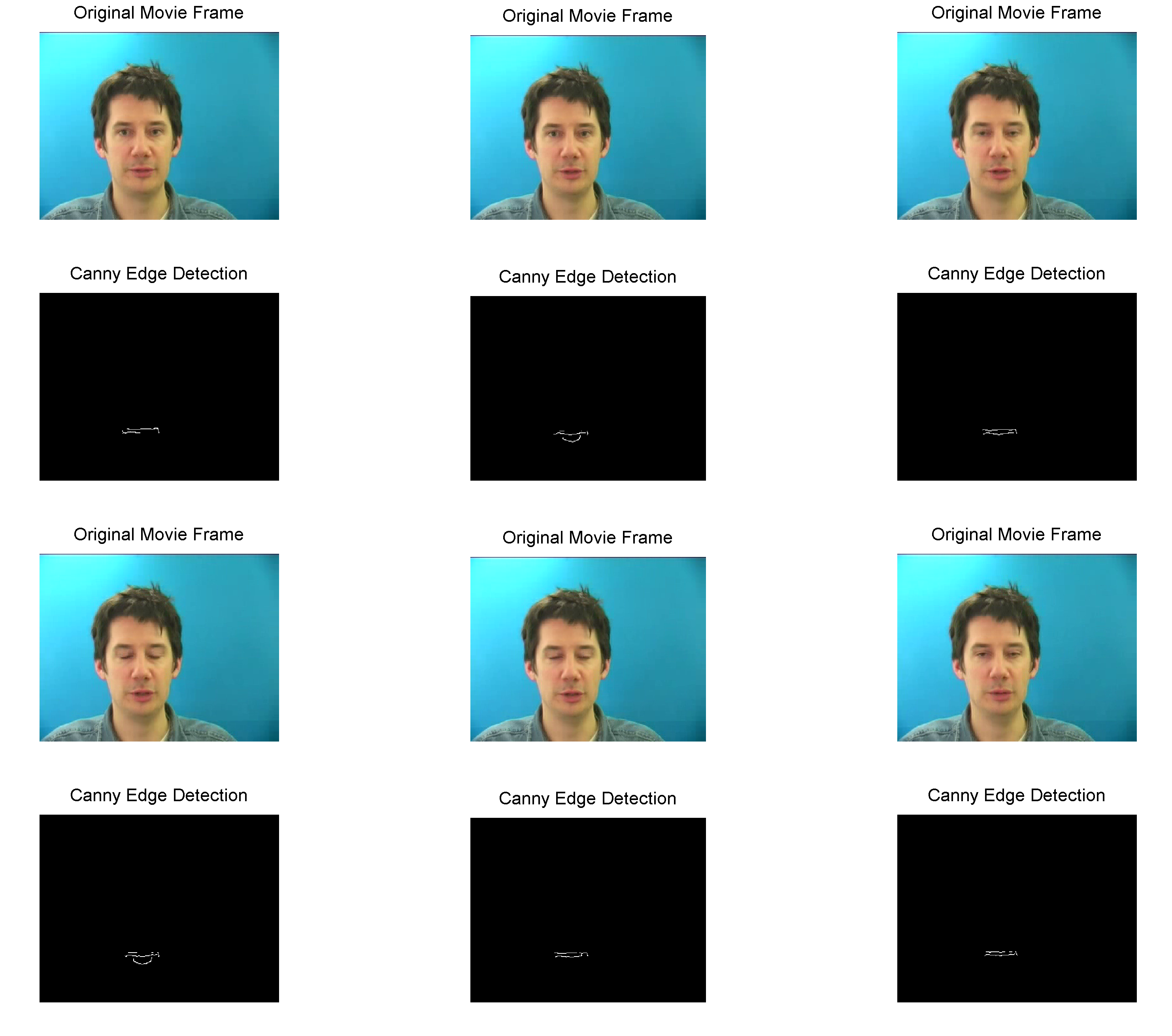}
	\caption{Canny Edge Detection results for selected frames. Although it captures the profile of the lips, it does not always capture the entire contour.It also does not show the inside of the lips.}
\end{figure} 

\subsubsection{Dynamic mode decomposition}
Next we use the DMD algorithm we developed in homework 4 to try foreground/background separation. We find that the face moves too much, and there is too much variation in the speaker's skin around the mouth to properly use DMD. We detect many points that are not on the lip. While DMD is good at separating the man's face from the background, it cannot accurately separate the lips from mouth region. It might be possible to alter the brightness and contrast of the separated image using DMD to obtain a better result, but it would still be too inconsistent to obtain good results for the rest of the project. \par

\begin{figure}[!ht]
	\includegraphics[width=1\textwidth, height=0.75\textwidth]{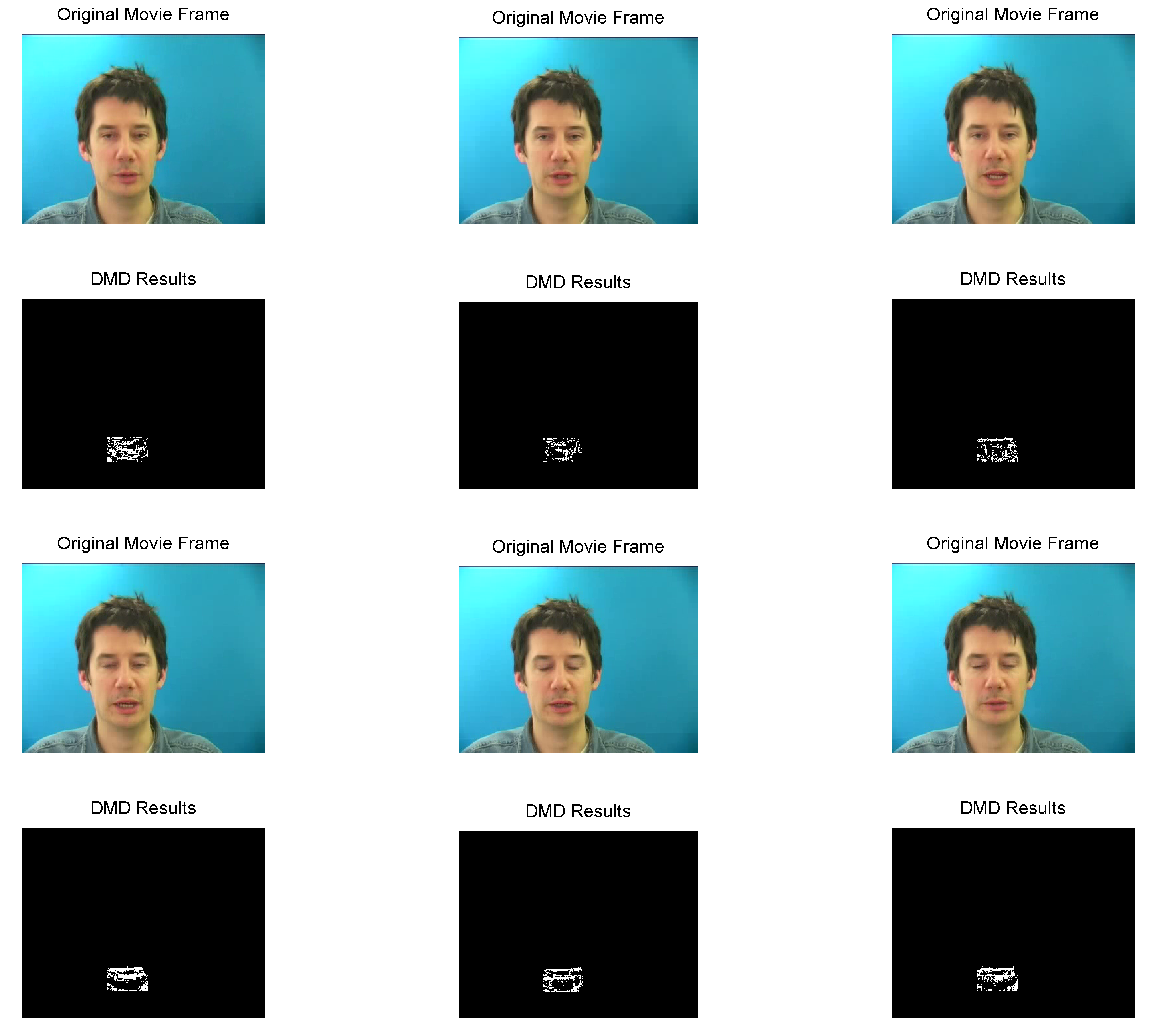}
	\caption{DMD results for selected frames. DMD is not ideal in this situation because the background (the man's face) is not completely stationary.}
\end{figure}
\subsubsection{Color classification}
Thus far, all of the methods discussed have acted on grayscale frames of the color movie. We now consider a technique for acting on the color frames. If we assume that the lips of the speaker have a distinct color from the rest of the speaker's mouth region (i.e. the skin, teeth, inside of the mouth) then we can determine the lip region by classifying the colors of the mouth region. It is possible to do this using the RGB values of each pixel. We instead represent each pixel with it's LAB color space values. The point of LAB colors  is that they match how human vision perceives colors. The L stands for lighting and AB are values corresponding to color. By classifying each pixel into one of up to 4 clusters based on it's AB values with a k-means algorithm, we can separate the lips from the rest of the mouth region. The number of clusters created varies per frame because in some frames, the teeth and inner mouth can be grouped in their own clusters. Other frames, the speaker has their mouth closed so only 2 clusters are needed. We can differentiate between the clusters by looking at the calculated centroid positions of each cluster. It is determined experimentally that the lip region has the highest A color value, and it usually has the lowest B color value as well.  Color classification is fairly fast and accurate. It gives the thickness as well as the shape of the lips. It's only downside is that it often classifies the left and right sides of the lips into a different cluster. Color classification typically picks out an ellipse in the middle of the lips as opposed to the whole lip region. Overall, color classification is determined to give the most accurate results, and edge detection is a close second. In hindsight, it would have made much more sense to use a classification algorithm, like knn-search, instead of k-means. Although we choose the wrong algorithm to use our results are still acceptable. \par
 
\begin{figure}[!ht]
	\includegraphics[width=1\textwidth, height=0.75\textwidth]{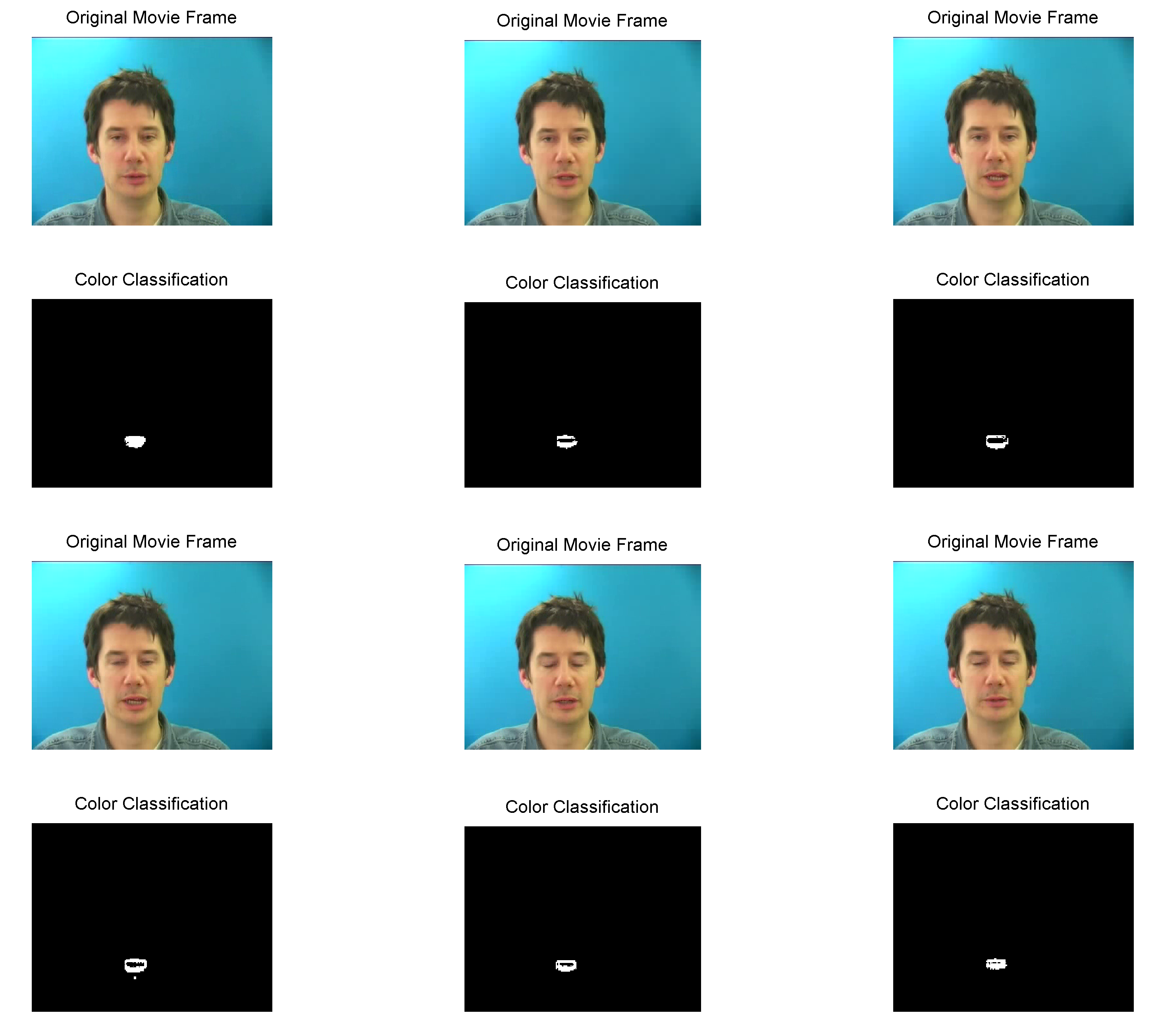}
 \caption{Color Classification for selected frames. Arguably the best overall method for capturing the movement of the lips.}
\end{figure}

In general the best results come from either more computationally expensive or mathematically sophisticated techniques. For example, training a neural net to classify the lip region of each frame would probably be the most effective technique overall, but would be very computationally costly and require manually determining the lip region for training. Many papers in the literature consider fitting a vector to fit the lip region and then track the lips by moving points of the vector and measuring the change of shape. They then can compute the most likely shape by comparing it to known positions of the lip. This method relies more heavily on analysis but also requires manual training as well. In general, lip tracking is a classification problem where the goal is to classify the pixels that belong to the lip of the speaker. 

\subsection{Extracting Phonemes}

Using the nltk library in Python, we convert every word to its constitutive arpabet phonetics. It gives the following output for the words - f', 'see', 'sea', 'compute', 'comput', 'cat'. Only 'comput' fails because it isn't a real word
\begin{verbatim}
['EH1', 'F']
['S', 'IY1']
['S', 'IY1']
['K', 'AH0', 'M', 'P', 'Y', 'UW1', 'T']
'comput'
['K', 'AE1', 'T']
\end{verbatim}

From the words spoken in our videos, we get a set of 36 unique phonemes. The code for this is shown in using \ref{Ph.py}. From the transcripts, we extract all the data into a csv file using \ref{Transcripts.py}

\subsection{Extracting Subtitles and Assigning Phonemes}
The Corpus Grid II database contains a {\it .txt} file for each video containing the words spoken and corresponding frames for each word.  These were downloaded and extracted into Matlab using the {\it textscan} function in Matlab.  The words were deconstructed into their phonemes, and phonemes were assigned to each frame to create the training set.  The assignment of a phoneme label to each frame was done by assigning each phoneme from a word to an equivalent number of the video frames corresponding to the phonemes in each word. 

\subsection{Phoneme and Viseme Classification}
The data matrix was created by reshaping each frame in each video in to a single column.  When each column was reshaped, its saved phoneme was checked, and the corresponding viseme index was saved to create the labels for a classification algorithm. The columns for each video frame were then concatenated in order to form the data matrix.  The singular value decomposition was computed of the matrix of video frames.  Classification was performed on first 30 columns of the $V$ matrix using both a Naive Bayes and a k-nearest neighbors algorithm.  Classification was done with a random $75\%$ of the data used for training and the remaining $25\%$ of the data for cross validation.

\subsection{Word Classification using HMMs}

Once a phoneme classifier has been created, we apply it onto the whole dataset of 1000 videos to obtain 1000 phoneme sequences of length 74. If we use phonemes, we will get sequences of numbers which are between 1 and 37. If we use visemes, we get sequences with numbers between 1 and 11.

Our objective is to map sequence of phonemes/visemes to words. This is where we utilize the capability of Hidden Markov Models to incorporate the temporal dynamics of the phoneme changes and thus "remember". So, even if our phoneme classification is erroneous, the patterns generated by the classifier can be mapped onto words resulting in higher accuracy.

We have the 1000 sequences.From the video dataset, \ref{binsorter.m} extracts sequences corresponding to each word (using regular expressions).
This leaves us with a list of sequences for each word. For each word, the list is divided up into training and test sets. We train an HMM for each word. This means that each word has a unique transition matrix and observation matrix associated with it. All the training is done in \ref{makethehmm2.m}. The functions used for training the HMM are from Kevin Murphy's HMM toolbox \cite{key-9}.

Then, when we have an unknown sequence, we run it through the HMMs of all the words and find the "loglikelihood" that it might be that word. When we find the HMM for which it has the highest "loglikelihood", it means that the sequence is most likely to be the word associated with that HMM. The results are estimated using \ref{resultshmm.m}.

\section{Computational Results}

\subsection{ Phoneme and Viseme Classification}
For phoneme identification, the classification using a k-nearest neighbors algorithm only $11.6103\%$ accuracy on average on the cross validation over 176 trials.  For viseme identification, the k-nearest neighbors method had $19.7355\%$ accuracy over 30 trials.  More trials were not performed due to the Matlab {\it knnsearch} function being computationally intensive.  

The naive bayes classification algorithm applied to identificaiton of phonemes had an average accuracy of $3.49\%$ over 1000 trials.  For viseme classification, the naive bayes algorithm had an average accuracy of $9.1357\%$ over 1000 trials. 

\subsection{Word Classification using HMMs}

The identification of the right word definitely depends on the words it is compared against. For this purpose, the results are presented in \ref{tab:phoneme} and \ref{tab:viseme} show the word to be detected, the other words and its accuracy.

Using the phoneme classification, the results seem to be good in general. Although the accuracy does go down with more words, it is still very good and definitely better than random guessing in most cases.

\begin{table}[]
	\caption{Classifying words after classifying frames into phonemes}
\begin{center}
\begin{tabular}{ |l|l|l| }
	\hline
	\multicolumn{3}{ |c| }{Word Classification using 37 phonemes} \\
	\hline
	Word & Set & Accuracy\\ \hline
	bin & bin , blue & 87.5 \% \\
	blue & bin, blue &  36 \% \\
	blue & red, blue & 76 \% \\
	four & four, white & 60 \% \\
	bin & bin , white & 62.5 \%  \\
	five & blue , five & 60 \%  \\
	red & red , eight & 72 \% \\ \hline
bin & bin , blue, green & 75 \% \\
green &  green, white, five & 44 \% \\ \hline
five &  five, blue, four, white & 50 \%  \\
green &  green, white, five, red & 28 \%  \\
bin & bin , blue, green , red & 75 \% \\
bin & bin , blue, red, white & 56.2 \% \\
blue & bin , blue, red, white & 28 \% \\	
five & four , five, red, white & 45 \% \\
\hline
bin & bin , blue, green , red, eight & 75 \% \\	
bin & bin , blue, green , red, white & 50 \% \\	
four & four , five, green , red, white & 30 \% \\
five & four , five, green , red, white & 40 \% \\	
	\hline
\end{tabular}
\end{center} 
\label{tab:phoneme}
\end{table}

We notice that the best identified words by the viseme classification might not be the same as those by the phoneme classification. The word "bin" is an example which is easily identified by the phoneme classification but not so well when the states are visemes. In contrast, the word "blue" is better identified now. On exploring the viseme assignment, the viseme 11 was quite over-assigned which might have played a role. The results are still very good for the viseme classification. 

\begin{table}[]
	\caption{Classifying words after classifying frames into visemes}
\begin{center}
	\begin{tabular}{ |l|l|l| }
		\hline
		\multicolumn{3}{ |c| }{Word Classification using 11 Visemes} \\
		\hline
		Word & Set & Accuracy\\ \hline
		bin & bin , blue & 50 \% \\
		blue & bin, blue &  68 \% \\
		blue & red, blue & 92 \% \\
		four & four, white & 75 \% \\
		bin & bin , white & 68.8 \%  \\
		five & blue , five & 65 \%  \\
		red & red , eight & 65 \% \\ \hline
		bin & bin , blue, green & 25 \% \\
		four & bin , four, green & 35 \% \\
		red & red , white, green & 35 \% \\
		green &  green, white, five & 32 \% \\
		green &  green, white, blue & 36 \% \\ \hline
		five &  five, blue, four, white & 50 \%  \\
		green &  green, white, five, red & 32 \%  \\
		bin & bin , blue, green , red & 12.5 \% \\
		bin & bin , blue, red, white & 12.5\% \\
		blue & bin , blue, red, white & 68 \% \\	
		five & four , five, red, white & 50 \% \\
		\hline
		bin & bin , blue, green , red, eight & 12.5 \% \\	
		bin & bin , blue, green , red, white & 12.5 \% \\	
		four & four , five, green , red, white & 30 \% \\
		five & four , five, green , red, white & 25 \% \\	
		\hline
	\end{tabular}
\end{center}
\label{tab:viseme}
\end{table}

Although there were a few more words and many more permutations to try out, we restricted ourselves to these selected ones under the time constraints.
\newpage

\section{Summary and Conclusions}

Finding bottlenecks during our attempt enables others to devise ways to evade them next time. There are a number of things we could improve upon.
\begin{itemize}
	\item The best way to improve the results in this project is to improve the lip detection.  The videos used in this project were of low quality, making the accurate detection of lip contours difficult.  The database used has higher quality versions of the videos. However high quality videos are $\approx 2.4$ GB each, for a total of 2.4 Terabytes of data for the whole data set.  This is an unrealistic amount of data for the hardware available for this project.
	\item Another short coming of this project is the assignment of an equal number of video frames to each phoneme from the given frame locations of each word.  This is a poor way to determine which phoneme is being spoken in each video frame. If we had a dataset in which we had frames corresponding to each phoneme, the accuracy would easily increase.
	\item In the phoneme classification, we used only naive bayes and k-nearest neighbours. Even though, knn gave us high accuracy, it was computationally expensive and we did not have sufficient time to classify the whole dataset. Results could have further been improved with a more sophisticated classification algorithm, for instance a classification tree or neural network.
	\item Apart from Hidden Markov Models, Recurrent Neural Networks are widely popular for activities that involve sequential events and memory. TensorFlow has a great RNN package in Python which is worth exploring.
	\item Kevin Murphy's HMM toolbox is widely popular as the best HMM toolbox in Matlab. However, it cannot handle sequences of different length during training. We worked around this problem by using stop states and chopping the sequence off. But that is far from a perfect fix. Searching for better toolboxes or making our own scripts to do the E-M step of the algorithm is a better idea. 
	\item We used 37 phonemes in our classification. That is a tall order for any classifier. By using 11 visemes, we tried to cut this down. This can be better improved choosing the sets to group together. Unsupervised learning could be a good option to experiment on to find good clusters.
	
	\item HMMs are trained using an initial guess of the transition and observation matrices. If we could figure out a way to get a good initial guess especially in the context of phonemes/visemes in speech, there would be a huge improvement.
	
	\item Lastly, 1000 videos would correspond to around 200 instances for the average word. This is not a particularly large sample size. So, using more videos will lead to better results.		
\end{itemize}

Given an unknown video sequence where only the words in our dictionary are spoken, we have a technique to identify the words spoken to a good level of accuracy. We also have a good exposure to top tier methods for image processing, classification and hidden markov models as well as their limitations. The project was also a good example of collaborative research and use modern day tools like Git for version control and to speed up progress. All the code, the report and some very useful .mat files can be found on our GitHub repository \cite{key-12}.



\newpage
\begin{thebibliography}{1}
	\bibitem{key-1}Davis, Abe, et al. "The visual microphone: passive recovery of sound from video." (2014)..
	\bibitem{key-2}Rabiner, Lawrence R. "A tutorial on hidden Markov models and selected applications in speech recognition." Proceedings of the IEEE 77.2 (1989): 257-286.	
	\bibitem{key-3}Choi, Kyoung Ho, and Jenq-Neng Hwang. "Baum-welch hidden Markov model inversion for reliable audio-to-visual conversion." Multimedia Signal Processing, 1999 IEEE 3rd Workshop on. IEEE, 1999.
	
	\bibitem{key-4}Yang, Jie, and Yangsheng Xu. Hidden markov model for gesture recognition. No. CMU-RI-TR-94-10. CARNEGIE-MELLON UNIV PITTSBURGH PA ROBOTICS INST, 1994.

	\bibitem{key-5}Starner, Thad E. Visual Recognition of American Sign Language Using Hidden Markov Models. MASSACHUSETTS INST OF TECH CAMBRIDGE DEPT OF BRAIN AND COGNITIVE SCIENCES, 1995.
	\bibitem{key-6}Hassanat, Ahmad B., `Visual Words for Automatic Lip- Reading.' PhD diss., University of Buckingham, 2009.
	\bibitem{key-7}J. Zhong, W. Chou, and E. Petajan, `Acoustic Driven Viseme Identification for Face Animation.' Bell Laboratories. Murray Hill, NJ. IEEE 0-7803-378. Aug. 1997.
	\bibitem{key-8}L. Cappelletta and N. Harte. `Phoneme-to-Viseme Mapping for Visual Speech.' Department of Electronic and Electrical Engineering, Trinity College Dublin, Ireland. May 2012.
	
	\bibitem{key-9}\url{https://www.cs.ubc.ca/~murphyk/Software/HMM/hmm.html}
	\bibitem{key-10}\url{http://spandh.dcs.shef.ac.uk/gridcorpus/}
	\bibitem{key-11}\url{http://www.ee.surrey.ac.uk/Projects/LILiR/datasets.html}
	\bibitem{key-12}\url{https://github.com/Dirivian/dynamic_lips}
\end{thebibliography}
\end{document}